\documentclass[10pt,twocolumn,letterpaper]{article}
\usepackage[final]{cvpr}

\usepackage[pagebackref,breaklinks,colorlinks]{hyperref}
\usepackage{svg}
\usepackage{times}
\usepackage{epsfig}
\usepackage{graphicx}
\usepackage{amsmath}
\usepackage{amssymb}

\graphicspath{figures}
\usepackage{tabularx}
\usepackage{booktabs}
\usepackage{dcolumn}
\usepackage{hyperref}
\usepackage{amsfonts}
\usepackage{subcaption}
\usepackage{threeparttable}
\usepackage{multirow}
\usepackage{mathrsfs}

\begin{document}
\title{View Transition based Dual Camera Image Fusion}
\author{Tiantian Cao\footnotemark[1], Xuan Dong\footnotemark[1], Chunli Peng, Zhengqing Li, Xinyu Guo, Weixin Li}

\maketitle
\renewcommand{\thefootnote}{\fnsymbol{footnote}} 
\footnotetext[1]{equal contribution.} 
\begin{abstract}
The dual camera system of wide-angle ({\bf{W}}) and telephoto ({\bf{T}}) cameras has been widely adopted by popular phones. In the overlap region, fusing the {\bf{W}} and {\bf{T}} images can generate a higher quality image. Related works perform pixel-level motion alignment or high-dimensional feature alignment of the {\bf{T}} image to the view of the {\bf{W}} image and then perform image/feature fusion, but the enhancement in occlusion area is ill-posed and can hardly utilize data from {\bf{T}} images. Our insight is to minimize the occlusion area and thus maximize the use of pixels from {\bf{T}} images. Instead of insisting on
placing the output in the {\bf{W}} view, we propose a view transition method to transform both {\bf{W}} and {\bf{T}} images into a mixed view and then blend them into the output. The transformation ratio is kept small and not apparent to users, and the center area of the output, which has accumulated a sufficient amount of transformation, can directly use the contents from the {\bf{T}} view to minimize occlusions. Experimental results show that, in comparison with the SOTA methods, occlusion area is largely reduced by our method and thus more pixels of the {\bf{T}} image can be used for improving the quality of the output image.
\end{abstract}

\section{Introduction}
\begin{figure}
\vspace{-0pt}
\centering
\begin{minipage}[ht]{.99\linewidth}
\centering
\subfloat[Wide-angle ({\bf{W}}) input]{\label{}\includegraphics[width=0.31\linewidth]{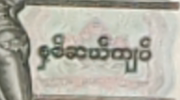}}\hspace{4pt}
\subfloat[Telephoto ({\bf{T}}) input]{\label{}\includegraphics[width=0.31\linewidth]{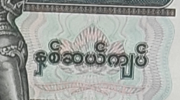}}\hspace{4pt}
\subfloat[Our output ({\bf{O}})]{\label{}\includegraphics[width=0.31\linewidth]{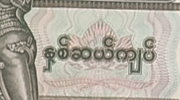}}\hspace{4pt}
\end{minipage}

\vspace{-10pt}

\caption{Example patches of {\bf{W}} and {\bf{T}} images to show their quality differences. This motivates us to fuse {\bf{T}} pixels into {\bf{W}} images.}
\label{fig:1}
\end{figure}

The heterogeneous dual camera system of a wide-angle camera and a telephoto camera, denoted as {\bf{W}} and {\bf{T}} respectively throughout this paper, has been widely adopted by the majority of phone vendors (In this paper, we assume the {\bf{T}} camera is on the left view, and the {\bf{W}} camera is on the right view). When capturing distant scenes, the system can simultaneously activate the {\bf{W}} camera and the {\bf{T}} camera for shooting. The {\bf{W}} image has wide field of view (FoV) and the {\bf{T}} image has significantly higher quality, as shown in Fig. \ref{fig:1}. It offers a new approach to produce high-quality images by fusing them.

\captionsetup[subfloat]{labelsep=none,format=plain,labelformat=empty}

\captionsetup[subfloat]{labelsep=none,format=plain,labelformat=empty}
\begin{figure}
\vspace{-10pt}
\centering
\begin{minipage}[ht]{.99\linewidth}
\centering
\subfloat[Wide-angle ({\bf{W}}) input]{\label{}\includegraphics[width=0.31 \linewidth]{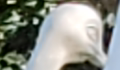}}\hspace{4pt}
\subfloat[Telephoto ({\bf{T}}) input]{\label{}\includegraphics[width=0.31\linewidth]{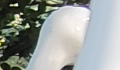}}\hspace{4pt}
\subfloat[Our output ({\bf{O}})]{\label{}\includegraphics[width=0.31\linewidth]{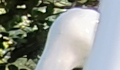}}\hspace{4pt}
\end{minipage}

\vspace{-10pt}

\caption{Example patches to show, since {\bf{W}} and {\bf{T}} cameras are at different views, there will be occlusions when objects have different depths. In these regions, the fusion becomes difficult.}
\label{fig:2}
\end{figure}

In the literature, one strategy uses pixel-level alignment followed by image fusion, e.g. \cite{lai2022face}. This strategy can effectively fuse {\bf{T}} image information into the final result in non-occluded regions. Another strategy is based on deep learning super-resolution methods, e.g. \cite{dcsr}. These methods often use high-dimensional aligned {\bf{T}} features as guidance to enhance the quality of {\bf{W}} image. Since the above two strategies still position the output, denoted as {\bf{O}} throughout this paper, image within the right view of the {\bf{W}} image, due to different views of {\bf{W}} and {\bf{T}} cameras, there exist many occlusion areas when using the {\bf{T}} image/feature to enhance the {\bf{W}} image, as shown in Fig. \ref{fig:2}. The assistance of the high-quality {\bf{T}} image for enhancing occluded regions is limited, resulting in a relatively low upper limit for enhancement in these areas.

\begin{figure}[h]
\begin{center}
   \includegraphics[width=0.99\linewidth]{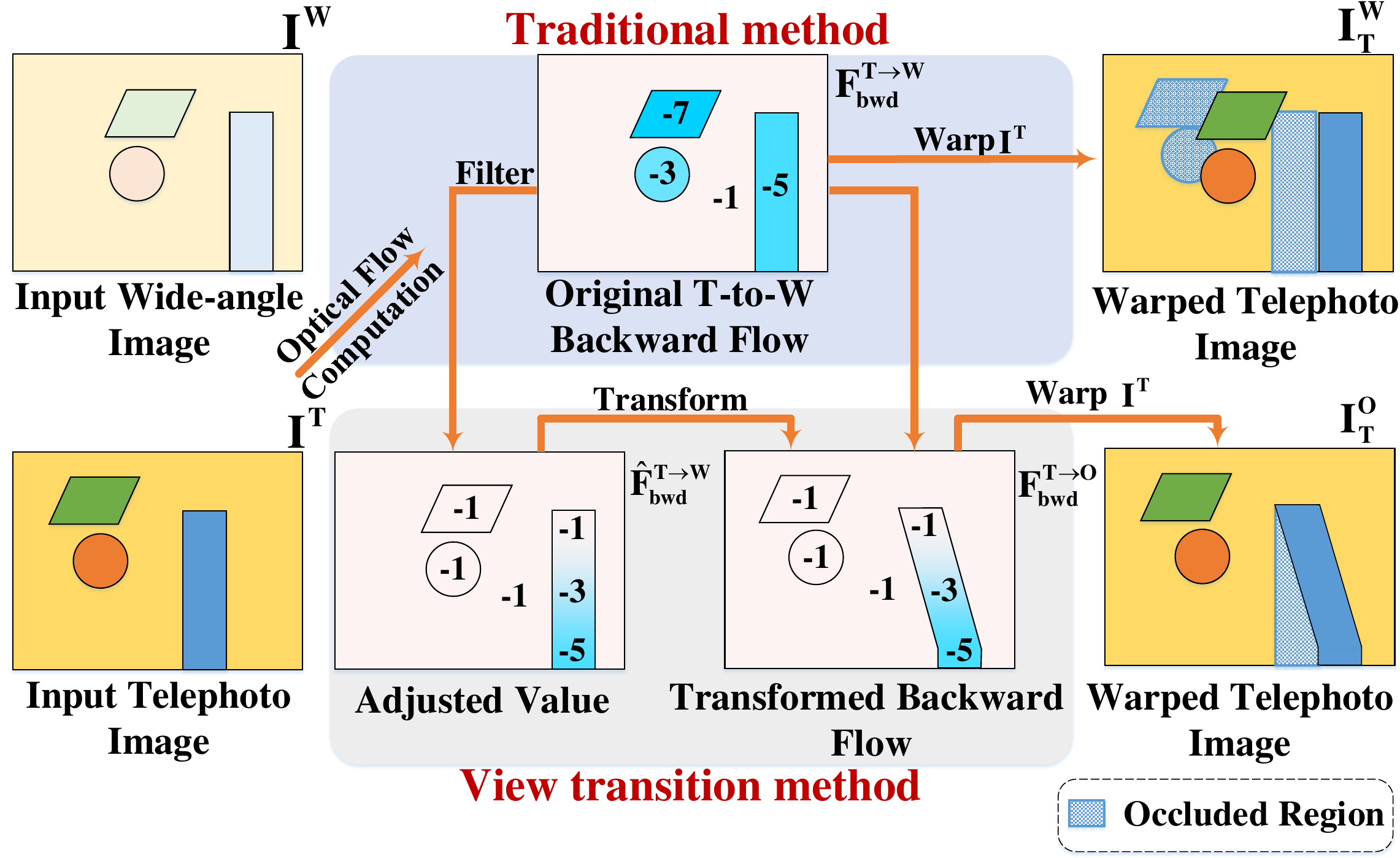}
\end{center}
\vspace{-10pt}
\caption{Our insight. Given input images ${\bf{I}}^{{\bf{W}}}$ and ${\bf{I}}^{{\bf{T}}}$, traditional methods usually compute the original {\bf{T}}-to-{\bf{W}} backward optical flow ${\bf{F}}_{bwd}^{{\bf{T}} \to {\bf{W}}}$ and use it to warp ${\bf{I}}^{{\bf{T}}}$ to get the warped result ${\bf{I}}_{\bf{T}}^{{\bf{W}}}$ for enhancing ${\bf{I}}^{{\bf{W}}}$, but there exist many occlusions in ${\bf{I}}_{\bf{T}}^{{\bf{W}}}$ where the enhancement is difficult. To minimize occlusions and ensure imperceptible transformations to users, our view transition strategy calculates a transformed flow ${\bf{F}}_{bwd}^{{\bf{T}} \to {\bf{O}}}$ and use it to warp ${\bf{I}}^{{\bf{T}}}$ to get the warped result ${\bf{I}}_{\bf{T}}^{{\bf{O}}}$. The view transition steps includes 1) adjusting the value of each pixel in ${\bf{F}}_{bwd}^{{\bf{T}} \to {\bf{W}}}$ to get ${\bf{\hat F}}_{bwd}^{{\bf{T}} \to {\bf{W}}}$ so that, from the boundary to the center, the flow values slowly transition to spatially smooth values, 2) revising the coordinate of each pixel in ${\bf{\hat F}}_{bwd}^{{\bf{T}} \to {\bf{W}}}$ according to the differences of ${\bf{F}}_{bwd}^{{\bf{T}} \to {\bf{W}}}$ and ${\bf{\hat F}}_{bwd}^{{\bf{T}} \to {\bf{W}}}$. As shown, the occlusions in ${\bf{I}}_{\bf{T}}^{{\bf{O}}}$ are much reduced than ${\bf{I}}_{\bf{T}}^{{\bf{W}}}$.}

\label{fig:insight}

\end{figure}

This paper focuses on the fusion of the overlap region of  {\bf{W}} and {\bf{T}} images. Our goal is to minimize the occlusion area, which lacks a high-quality reference and is difficult to enhance, and thus maximize the usage of pixels from {\bf{T}} images to improve the quality of {\bf{O}} images. Because {\bf{W}} and {\bf{T}} cameras on smartphones are closely arranged, and the difference in view is not significant, users are not particularly concerned whether the {\bf{O}} content is from the right view or the left view. Therefore, our insight is that, as shown in Fig. \ref{fig:insight}, we don't insist on placing the {\bf{O}} image in the right view. Instead, we place it in a mixed view between the left and right views. Specifically, in the central area of the {\bf{O}} image, we utilize the left view of the {\bf{T}} image, minimizing occlusion and maximizing the use of {\bf{T}} pixels. In the boundary regions, we utilize the right view of the {\bf{W}} image to maintain consistency with the content of the {\bf{W}} image outside the overlap region. In the transitional zone from the boundary to the center, we gradually transition from the right view to the left view. The shape change of image content during the transition is kept small enough and not apparent to users.

\begin{figure}[h]
\begin{center}
   \includegraphics[width=0.99\linewidth]{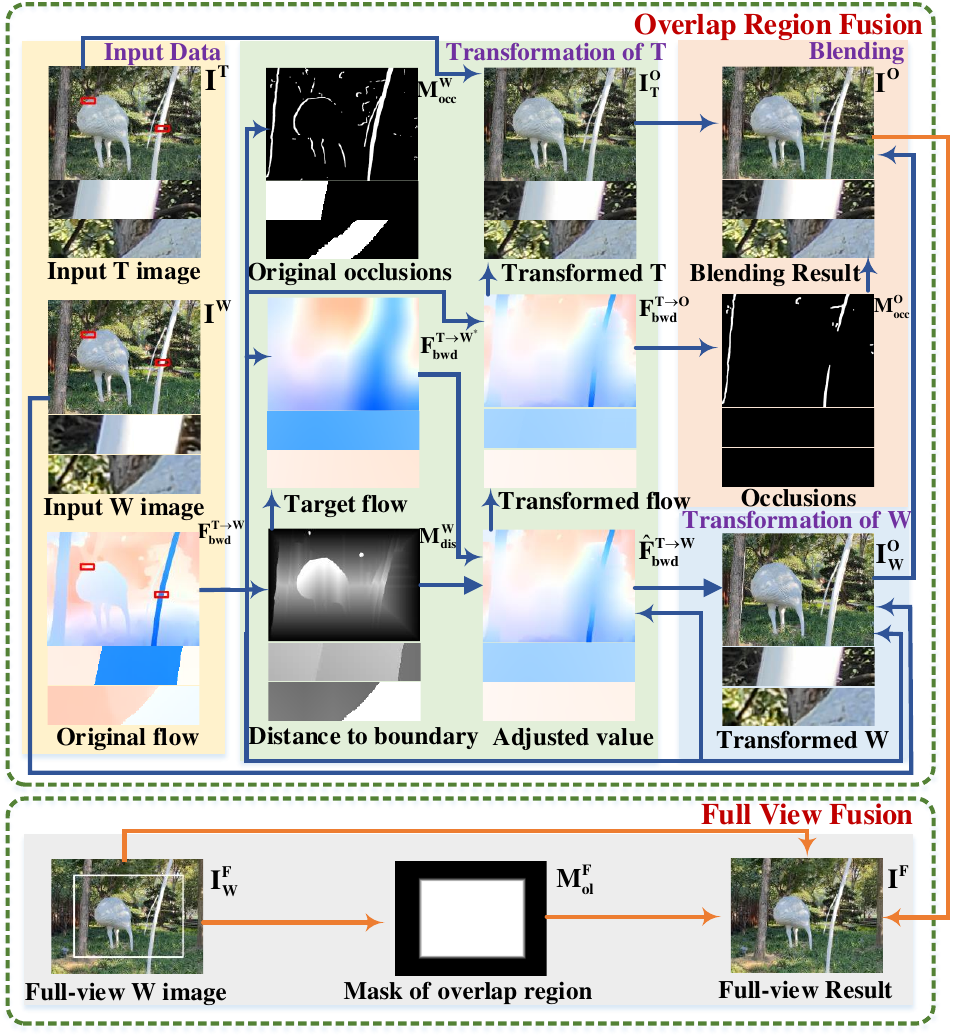}
\end{center}

    \vspace{-10pt}

   \caption{Pipeline of our method. In the core overlap region fusion part, the input includes images ${\bf{I}}^{{\bf{T}}}$ and ${\bf{I}}^{{\bf{W}}}$ and the original {\bf{T}}-to-{\bf{W}} backward flow ${\bf{F}}_{bwd}^{{\bf{T}} \to {\bf{W}}}$. 1) We compute the required transformed flow ${\bf{F}}_{bwd}^{{\bf{T}} \to {\bf{O}}}$ to warp ${\bf{I}}^{{\bf{T}}}$ to the ${\bf{O}}$ view result ${\bf{I}}_{\bf{T}}^{{\bf{O}}}$. 2) We transform ${\bf{I}}^{{\bf{W}}}$ to the ${\bf{O}}$ view result ${\bf{I}}_{\bf{W}}^{{\bf{O}}}$ according to the transformation of ${\bf{I}}^{{\bf{T}}}$. 3) We blend ${\bf{I}}_{\bf{T}}^{{\bf{O}}}$ and ${\bf{I}}_{\bf{W}}^{{\bf{O}}}$ to get the overlap-region result ${\bf{I}}^{{\bf{O}}}$. In the full-view fusion part, we blend ${\bf{I}}^{{\bf{O}}}$ with the non-overlap regions of ${\bf{I}}_{\bf{W}}^{{\bf{F}}}$ to get the full-view result ${\bf{I}}^{{\bf{F}}}$. The regions marked in red boxes are enlarged.}
\vspace{-10pt}
\label{fig:IR}
\end{figure}

Since the view transition operation lacks explicit supervision rules or supervised data, we address this problem using traditional non-learning-based image processing methods. The view transition is fundamentally a non-rigid image transformation issue. We calculate a motion vector for each pixel, known as optical flow, and then use image warping to achieve the required view transition. The pipeline is shown in Fig.\ref{fig:IR}.  1) We revise the original {\bf{T}}-to-{\bf{W}} backward flow ${\bf{F}}_{bwd}^{{\bf{T}} \to {\bf{W}}}$ to obtain the required transformed backward flow ${\bf{F}}_{bwd}^{{\bf{T}} \to {\bf{O}}}$ for transforming the {\bf{T}} image to the {\bf{O}} view, with the goal of minimizing occlusions and ensuring imperceptible transformations to users. 2) The {\bf{W}} image is transformed to the {\bf{O}} view according to the transformation of the {\bf{T}} image. 3) We use region-based histogram equalization \cite{histogram} to reduce tone inconsistency between the transformed {\bf{T}} and {\bf{W}} images. Then, we use pyramid blending \cite{mertens2007exposure} to smoothly blend the non-occluded regions of the transformed {\bf{T}} image with the occluded regions of the transformed {\bf{W}} image and the non-overlap regions of the full-view {\bf{W}} image.

Experiments are conducted on CameraFusion \cite{dcsr} and OPPO72, a dataset that we collected, consisting of 72 pairs of {\bf{W}} and {\bf{T}} images captured by an OPPO cellphone. Benefiting from the view transition, the occlusion area is reduced from 12.71\% and 17.75\% with the traditional method to 4.27\% and 4.28\% with our method on the OPPO72 and CameraFusion respectively. Thus, our usage of pixels of the {\bf{T}} image in the overlap region is increased significantly, achieving superior quality than all comparison methods.

Contributions: the proposed view transition method effectively reduces occlusion areas, allowing for fusing more high-quality pixels from the {\bf{T}} image into the {\bf{W}} image, all while avoiding noticeable shape distortions.

\section{Related Work}
Dual-camera image fusion is usually solved by learning based super resolution methods. Different image/feature alignment techniques are studied, e.g. attention operations in \cite{dcsr}, feature warping in \cite{dzsr}, and optical flow in \cite{zedusr} \cite{lai2022face}. After alignment, various joint super-resolution models are proposed, e.g. CNN model \cite{dcsr}, ResNet \cite{dzsr}, RCAN \cite{zedusr}, and UNet \cite{lai2022face}. Supervised learning using synthetic training data is used in \cite{dcsr} \cite{lai2022face}. Self-supervised learning is also studied using self-supervised losses, e.g. \cite{dcsr}, and self-designed training data \cite{dzsr} \cite{zedusr}.

Reference-based super resolution methods are closely related to the problem in this paper. TTSR \cite{TTSR} employs cross-attention for high-dimensional feature alignment and combines results from single-image super-resolution with those from joint super-resolution. SRNTT \cite{SRNTT} utilizes patch-based alignment with patch feature inner products as a similarity measure and employs ResNet \cite{resnet} for joint super-resolution. In MASA \cite{masa}, alignment involves block-based search followed by pixel-level alignment. SAM normalization and a variant of a CNN are used for joint super-resolution. In Shim20 \cite{c2_matching}, alignment utilizes a U-Net network with non-local blocks to estimate offsets and employs Dconv \cite{dai2017deformable} for implementation. RCAN \cite{RCAN} is used for joint super-resolution.

All the above methods use the {\bf{T}} image as reference to enhance the {\bf{W}} image at the {\bf{W}} view. So, the enhancement of pixels in occluded regions is more like single image super-resolution. In occluded regions, the degradation model from the wide image to the high-quality image is unknown, and solving this problem is ill-posed. As a result, there still remains a significant quality gap between existing enhancement results and the real {\bf{T}} images.

Single image super resolution (SISR) is a general problem and can be used in different applications. The SOTA methods can be classified as CNN based \cite{RCAN}\cite{EDSR}, Transformer based \cite{SwinIR}, GAN based \cite{realesrgan}\cite{srgan} according to the deep model type. When performing SISR methods in enhancing the {\bf{W}} image, the highly correlated high-quality details in the {\bf{T}} image are not used, leading to the quality of the output image not competitive.

\section{Method}

This paper focuses on the fusion of the overlap region of the {\bf{W}} and {\bf{T}} images. As shown in the pipeline of Fig. \ref{fig:IR}, the main modules include the transformation of {\bf{T}} image, transformation of {\bf{W}} image, and image blending. The computational complexity of all our operations are $O(n)$ for efficient computation, except the optical flow computation.

Based on the insight in Fig. \ref{fig:insight}, the core of view transition is the transformation of the {\bf{T}} image. We revise the original {\bf{T}}-to-{\bf{W}} backward optical flow ${\bf{F}}_{bwd}^{{\bf{T}} \to {\bf{W}}}$ to obtain the required transformed flow ${\bf{F}}_{bwd}^{{\bf{T}} \to {\bf{O}}}$ which can warp ${\bf{I}}^{{\bf{T}}}$ to the {\bf{O}} view result ${\bf{I}}_{\bf{T}}^{{\bf{O}}}$. The goal of ${\bf{F}}_{bwd}^{{\bf{T}} \to {\bf{O}}}$ includes that 1) it can minimize the occluded areas of the transformed image ${\bf{I}}_{\bf{T}}^{{\bf{O}}}$. This entails minimizing the spatially discontinuous values in ${\bf{F}}_{bwd}^{{\bf{T}} \to {\bf{O}}}$. In addition, 2) ${\bf{F}}_{bwd}^{{\bf{T}} \to {\bf{O}}}$ should ensure no obvious shape changes of contents in ${\bf{I}}_{\bf{T}}^{{\bf{O}}}$. This entails that the variation from ${\bf{F}}_{bwd}^{{\bf{T}} \to {\bf{W}}}$ to ${\bf{F}}_{bwd}^{{\bf{T}} \to {\bf{O}}}$ needs to be constrained by the pixel's distance to the boundary, where the closer to the boundary, the smaller the degree of change. We propose that the variations should be smaller than the threshold permitted by radial distortion \cite{smith1992correction}. 

${\bf{I}}^{{\bf{W}}}$ is transformed to the {\bf{O}} view result ${\bf{I}}_{\bf{W}}^{{\bf{O}}}$ according to the transformation of the {\bf{T}} image. We use forward warp to perform the transformation. To overcome the artifacts, e.g. discontinuous lines, caused by quantization errors in forward warping, we apply minor adjustments to the forward optical flow with multiple times and average the results of multiple warps to get ${\bf{I}}_{\bf{W}}^{{\bf{O}}}$.

In the blending part, discrepancies in image tone exist between the ${\bf{I}}_{\bf{T}}^{{\bf{O}}}$ and ${\bf{I}}_{\bf{W}}^{{\bf{O}}}$ due to variations in the imaging pipeline and parameters of the {\bf{T}} and {\bf{W}} cameras. Classical global method \cite{Reinhard} and local method of Poisson blending \cite{poisson} are not good enough, as shown in Fig. \ref{fig:color}. We employ regional histogram equalization \cite{histogram} adjustments to mitigate the tone inconsistency.
In the blending operation, We employ classic pyramid blending \cite{mertens2007exposure}. The non-occluded areas utilize ${\bf{I}}_{\bf{T}}^{{\bf{O}}}$, and the occluded areas utilize ${\bf{I}}_{\bf{W}}^{{\bf{O}}}$.

Last, the overlap region fusion result ${\bf{I}}^{{\bf{O}}}$ is blended with the contents of ${\bf{I}}_{\bf{W}}^{{\bf{F}}}$ outside the overlap region via pyramid blending as well to generate the full-view result ${\bf{I}}^{{\bf{F}}}$.

\subsection{Transformation of {\bf{T}}}
The computation of original flow, target flow value, distance to boundary, and transformed flow are detailed below.

\begin{figure}
\vspace{-0pt}
\centering
\begin{minipage}[ht]{.99\linewidth}
\centering
\subfloat[]{\label{}\includegraphics[width=0.23\linewidth]{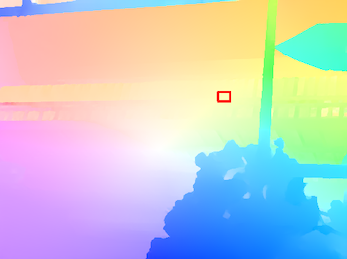}}\hspace{1pt}
\subfloat[]{\label{}\includegraphics[width=0.23\linewidth]{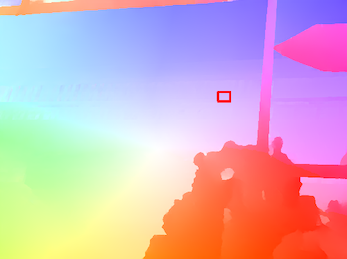}}\hspace{1pt}
\subfloat[]{\label{}\includegraphics[width=0.23\linewidth]{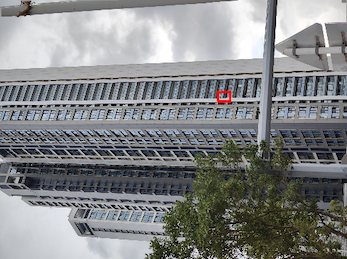}}\hspace{1pt}
\subfloat[]{\label{}\includegraphics[width=0.23\linewidth]{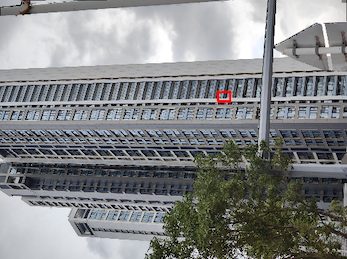}}\hspace{1pt}
\end{minipage}
\begin{minipage}[ht]{.99\linewidth}
\centering
\vspace{-12pt}
\subfloat[Backward flow]{\label{}\includegraphics[width=0.23\linewidth]{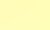}}\hspace{1pt}
\subfloat[Forward flow]
{\label{}\includegraphics[width=0.23\linewidth]{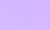}}\hspace{1pt}
\subfloat[Backward warp]{\label{}\includegraphics[width=0.23\linewidth]{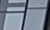}}\hspace{1pt}
\subfloat[Forward warp]{\label{}\includegraphics[width=0.23\linewidth]{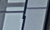}}\hspace{1pt}
\end{minipage}

\vspace{-10pt}

\caption{An example to compare backward warp with forward warp. This verifies our choice of backward warp to avoid artifacts.}
\label{fig:Wfwd_vs_Wbwd}
\end{figure}

\textbf{Original flow.} 
We use FlowFormer \cite{huang2022flowformer} as the optical flow method. We estimate the backward flow from the {\bf{T}} image to the {\bf{W}} image, i.e.
\begin{equation}
    {\bf{F}}_{bwd}^{{\bf{T}} \to {\bf{W}}} = FlowFormer({{\bf{I}}^{\bf{W}}},{{\bf{I}}^{\bf{T}}}),
\end{equation}
We estimate the backward flow instead of the forward flow because we aim to perform backward warp for the {\bf{T}} image which can generate more natural warping result than forward warp, as introduced in \cite{szeliski2022computer} and shown in Fig. \ref{fig:Wfwd_vs_Wbwd}.

\captionsetup[subfloat]{labelsep=none,format=plain,labelformat=empty}
\begin{figure}
\vspace{-0pt}
\centering
\begin{minipage}[ht]{.99\linewidth}
\centering
\subfloat[]{\label{}\includegraphics[width=0.23\linewidth]{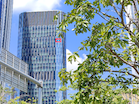}}\hspace{1pt}
\subfloat[]{\label{}\includegraphics[width=0.23\linewidth]{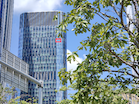}}\hspace{1pt}
\subfloat[]{\label{}\includegraphics[width=0.23\linewidth]{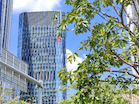}}\hspace{1pt}
\subfloat[]{\label{}\includegraphics[width=0.23\linewidth]{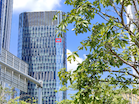}}\hspace{1pt}
\end{minipage}
\begin{minipage}[ht]{.99\linewidth}
\vspace{-12pt}
\centering
\subfloat[Input \textbf{W}]{\label{}\includegraphics[width=0.23\linewidth]{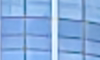}}\hspace{1pt}
\subfloat[Input \textbf{T}]{\label{}\includegraphics[width=0.23\linewidth]{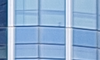}}\hspace{1pt}
\subfloat[$k=100 \times 100$]{\label{}\includegraphics[width=0.23\linewidth]{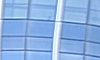}}\hspace{1pt}
\subfloat[$k=600 \times 600$]{\label{}\includegraphics[width=0.23\linewidth]{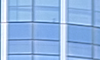}}\hspace{1pt}
\end{minipage}

\vspace{-10pt}

\caption{An example to show our results using different kernel sizes of the box filter. This verifies our choice of large kernel size ($k=600 \times 600$) to avoid obvious shape change of objects.}
\label{fig:box_filter}
\end{figure}

\textbf{Target flow value.} 
Values of all pixels in the target flow map should be spatially smooth, thereby reducing occlusions. First, we calculate the average flow ${{\bf{F}}_{\bf{M}}}$ from ${\bf{F}}_{bwd}^{{\bf{T}} \to {\bf{W}}}$ by box filter, i.e.
\begin{equation}
    {{\bf{F}}_{\bf{M}}} = BoxFilter({\bf{F}}_{bwd}^{{\bf{T}} \to {\bf{W}}},k),
\end{equation}
where the kernel $k=600 \times 600$ is set with a very large value to keep the transformation of contents that is determined by the flow values smooth enough. According to larger or smaller than ${{\bf{F}}_{\bf{M}}}$, each pixel in ${\bf{F}}_{bwd}^{{\bf{T}} \to {\bf{W}}}$ is judged as foreground or background, resulting in a binary mask image ${\bf{M}}$. Next, we compute the foreground flow ${{\bf{F}}_{\bf{F}}}$ by
\begin{equation}
    {{\bf{F}}_{\bf{F}}} = \frac{{BoxFilter({\bf{F}}_{bwd}^{{\bf{T}} \to {\bf{W}}} \cdot {\bf{M}},k)}}{{BoxFilter({\bf{M}},k)}}
\end{equation}
and the background flow ${{\bf{F}}_{\bf{B}}}$ by 
\begin{equation}
    {{\bf{F}}_{\bf{B}}} = \frac{{BoxFilter({\bf{F}}_{bwd}^{{\bf{T}} \to {\bf{W}}} \cdot (1-{\bf{M}}),k)}}{{BoxFilter((1-{\bf{M}}),k)}}.
\end{equation}
The target flow value for each pixel is proposed to be the average of foreground flow value and background flow value to reduce changes from ${\bf{F}}_{bwd}^{{\bf{T}} \to {\bf{W}}}$ so as to reduce shape change of contents. To make the target flow ${\bf{F}}_{bwd}^{{\bf{T}} \to {\bf{W}}*}$ smooth enough, we perform the box filter again to get ${\bf{F}}_{bwd}^{{\bf{T}} \to {\bf{W}}*}$, i.e.
\begin{equation}
    {\bf{F}}_{bwd}^{{\bf{T}} \to {\bf{W}}*} = BoxFilter(\frac{{{{\bf{F}}_{\bf{F}}} + {{\bf{F}}_{\bf{B}}}}}{2},k)
\end{equation}

We use the box filters above to take advantage of integral images for acceleration. As shown in Fig. \ref{fig:box_filter}, the kernel size $k$ should be large enough to avoid obvious shape change.

\begin{figure}
\vspace{-0pt}
\centering
\begin{minipage}[ht]{.99\linewidth}
\centering
\subfloat[(a) Input \textbf{W}]{\label{}\includegraphics[width=0.18\linewidth]{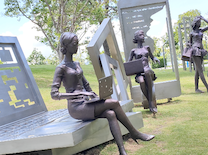}}\hspace{1pt}
\subfloat[(b) Input \textbf{T}]{\label{}\includegraphics[width=0.18\linewidth]{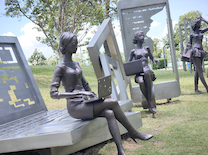}}\hspace{1pt}
\subfloat[(c) Base]{\label{}\includegraphics[width=0.18\linewidth]{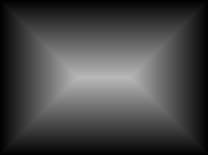}}\hspace{1pt}
\subfloat[(d) Our ]{\label{}\includegraphics[width=0.18\linewidth]{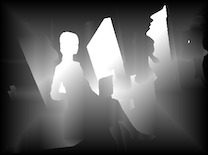}}\hspace{1pt}
\subfloat[(e) Distance]{\label{}\includegraphics[width=0.18\linewidth]{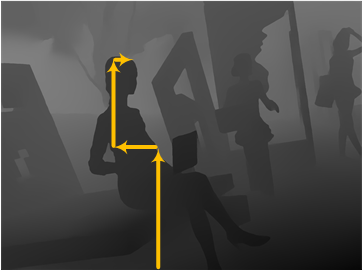}}\hspace{1pt}
\raggedleft \footnotesize distance \hspace{15pt}distance \hspace{8pt}computation 
\vspace{-8pt}
\end{minipage}

\caption{Distance to boundary maps using the baseline coordinate distance to boundary (c) and our method (d). The calculated distance values of our method is much larger than the baseline method in many areas, thereby increasing the transformation potential and reducing occlusions in these areas. In (e), we use the example pixel in the face region to explain how our method computes the distance of each pixel to the boundary considering the non-connected points with high gradient in the optical flow magnitude map.}
\label{fig:distance}
\end{figure}

\textbf{Distance to boundary.} 
One baseline method to compute how far each pixel is from the boundary is to calculate the shortest distance from a pixel's coordinates $(j, i)$ to each of the four boundaries. However, as shown in the absolution optical flow map of Fig. \ref{fig:distance} (e), some objects have relatively independent optical flow values, and the boundary that affects content coherence may not necessarily be the closest boundary to that pixel coordinate.

Based on the gradient of the optical flow map, we mark pixels with high gradient as non-connected points. In the presence of non-connected points, we calculate the shortest distance from each pixel to the four boundaries. An example to explain our method is shown in Fig. \ref{fig:distance} (e). The computed distance map is named ${\bf{M}}_{dis}^{\bf{W}}$. As shown in Fig. \ref{fig:distance} (c) (d), our result obtains larger distance values for independent objects so that they gain more potential for shape change and occlusion reduction.

\begin{figure}
\vspace{-0pt}
\centering
\begin{minipage}[ht]{.99\linewidth}
\centering
\subfloat[]{\label{}\includegraphics[width=0.23\linewidth]{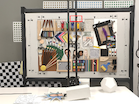}}\hspace{1pt}
\subfloat[]{\label{}\includegraphics[width=0.23\linewidth]{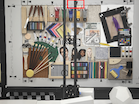}}\hspace{1pt}
\subfloat[]{\label{}\includegraphics[width=0.23\linewidth]{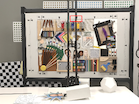}}\hspace{1pt}
\subfloat[]{\label{}\includegraphics[width=0.23\linewidth]{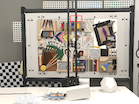}}\hspace{1pt}
\end{minipage}
\begin{minipage}[ht]{.99\linewidth}
\centering
\vspace{-12pt}
\subfloat[Input \textbf{W}]{\label{}\includegraphics[width=0.23\linewidth]{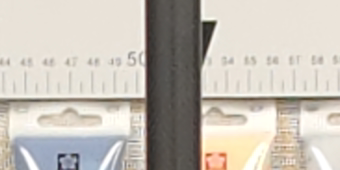}}\hspace{1pt}
\subfloat[Input \textbf{T}]{\label{}\includegraphics[width=0.23\linewidth]{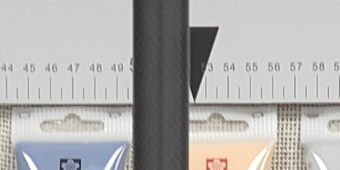}}\hspace{1pt}
\subfloat[ratio = 0.05]{\label{}\includegraphics[width=0.23\linewidth]{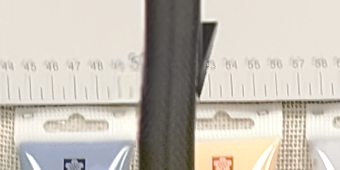}}\hspace{1pt}
\subfloat[ratio = 0.01]{\label{}\includegraphics[width=0.23\linewidth]{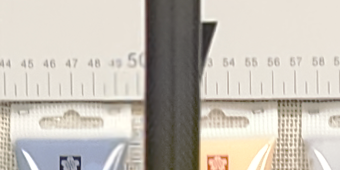}}\hspace{1pt}
\end{minipage}

\vspace{-10pt}

\caption{An example to show our results using different transformation ratio. The larger ratio value of 0.05 will cause a curvature of straight lines, while our choice of the smaller value of 0.01 can ensure the transformation not easily noticeable.}
\label{fig:ratio}
\end{figure}

\textbf{Transformed flow.} There are two steps to compute the transformed flow ${\bf{F}}_{bwd}^{{\bf{T}} \to {\bf{O}}}$, i.e. 1) adjusting the value of each pixel in ${\bf{F}}_{bwd}^{{\bf{T}} \to {\bf{W}}}$ to obtain ${\bf{\hat F}}_{bwd}^{{\bf{T}} \to {\bf{W}}}$, and 2) revising the coordinate of each pixel of ${\bf{\hat F}}_{bwd}^{{\bf{T}} \to {\bf{W}}}$ to get ${\bf{F}}_{bwd}^{{\bf{T}} \to {\bf{O}}}$.

In the first step, we notice that, in {\bf{W}} cameras, the coefficient of radial distortion is typically 0.01 \cite{distortion}. Therefore, we set the transformation ratio to 0.01 as well. Consequently, the permitted variation for ${\bf{F}}_{bwd}^{{\bf{T}} \to {\bf{W}}}$ is ${\bf{L}} = 0.01 \times {\bf{M}}_{dis}^{\bf{W}}$, i.e. the upper bound is ${\bf{F}}_{bwd}^{{\bf{T}} \to {\bf{W}}} + {\bf{L}}$ and lower bound is ${\bf{F}}_{bwd}^{{\bf{T}} \to {\bf{W}}} - {\bf{L}}$. We clip ${\bf{F}}_{bwd}^{{\bf{T}} \to {\bf{W*}}}$ by the upper and lower bounds to obtain ${\bf{\hat F}}_{bwd}^{{\bf{T}} \to {\bf{W}}}$, i.e.
\begin{equation}
    {\bf{\hat F}}_{bwd}^{{\bf{T}} \to {\bf{W}}} = clip({\bf{F}}_{bwd}^{{\bf{T}} \to {\bf{W}}*},{\bf{F}}_{bwd}^{{\bf{T}} \to {\bf{W}}} - {\bf{L}},{\bf{F}}_{bwd}^{{\bf{T}} \to {\bf{W}}} + {\bf{L}}).
\end{equation}

\begin{figure}
\vspace{-0pt}
\centering
\begin{minipage}[ht]{.99\linewidth}
\centering
\vspace{-0pt}
\subfloat[]{\label{}\includegraphics[width=0.23\linewidth]{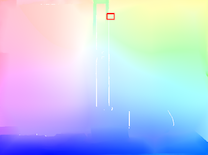}}\hspace{1pt}
\subfloat[]{\label{}\includegraphics[width=0.23\linewidth]{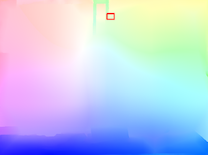}}\hspace{1pt}
\subfloat[]{\label{}\includegraphics[width=0.23\linewidth]{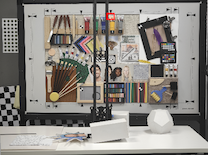}}\hspace{1pt}
\subfloat[]{\label{}\includegraphics[width=0.23\linewidth]{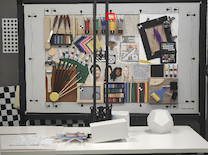}}\hspace{1pt}
\end{minipage}
\begin{minipage}[ht]{.99\linewidth}
\centering
\vspace{-12pt}
\subfloat[(a) Flow w/o]{\label{}\includegraphics[width=0.23\linewidth]{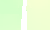}}\hspace{1pt}
\subfloat[(b) Our filled]
{\label{}\includegraphics[width=0.23\linewidth]{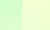}}\hspace{1pt}
\subfloat[Warped by (a)]{\label{}\includegraphics[width=0.23\linewidth]{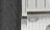}}\hspace{1pt}
\subfloat[Warped by (b)]{\label{}\includegraphics[width=0.23\linewidth]{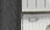}}\hspace{1pt}
\end{minipage}
 \footnotesize \hspace{10pt}filling \hspace{40pt} flow \hspace{110pt}

\vspace{-10pt}

\caption{Examples of the transformed flows without and with the empty region filling and the corresponding warped {\bf{T}} images using them.}
\label{fig:empty}
\end{figure}

In the second step, according to the flow changes between ${\bf{F}}_{bwd}^{{\bf{T}} \to {\bf{W}}}$ and ${\bf{\hat F}}_{bwd}^{{\bf{T}} \to {\bf{W}}}$, we change the coordinate of each pixel in ${\bf{\hat F}}_{bwd}^{{\bf{T}} \to {\bf{W}}}$ to their respective positions by forward warp, named ${Warp_{fwd}}$. After the forward warp process, the new optical flow map may contain some empty regions. The reason is that some areas that were originally occluded by foreground objects in the {\bf{W}} view become exposed after the transformation of contents to the {\bf{O}} view, as shown in Fig. \ref{fig:empty}. These regions should be filled with the optical flow values of the background. Therefore, we search for the optical flow values of the background objects near these empty regions and fill them with these values. The empty region filling operation is named $EF$. Thus, ${\bf{F}}_{bwd}^{{\bf{T}} \to {\bf{O}}}$ is computed by
\begin{equation}
    {\bf{F}}_{bwd}^{{\bf{T}} \to {\bf{O}}} = EF({Warp_{fwd}}({\bf{\hat F}}_{bwd}^{{\bf{T}} \to {\bf{W}}},{\bf{\hat F}}_{bwd}^{{\bf{T}} \to {\bf{W}}}-{\bf{F}}_{bwd}^{{\bf{T}} \to {\bf{W}}})).
    \label{eqn:Wfwd}
\end{equation}
${\bf{I}}_{\bf{T}}^{\bf{O}}$ is obtained by 
\begin{equation}
    {\bf{I}}_{\bf{T}}^{\bf{O}} = {Warp_{bwd}}({{\bf{I}}^{\bf{T}}},{\bf{F}}_{bwd}^{{\bf{T}} \to {\bf{O}}}).
\end{equation}

\begin{figure}
\vspace{-0pt}
\centering
\begin{minipage}[ht]{.99\linewidth}
\centering
\subfloat[]{\label{}\includegraphics[width=0.23\linewidth]{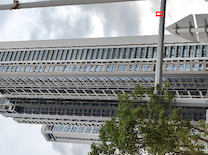}}\hspace{1pt}
\subfloat[]{\label{}\includegraphics[width=0.23\linewidth]{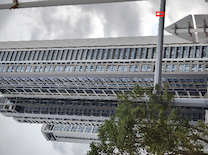}}\hspace{1pt}
\subfloat[]{\label{}\includegraphics[width=0.23\linewidth]{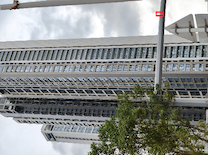}}\hspace{1pt}
\subfloat[]{\label{}\includegraphics[width=0.23\linewidth]{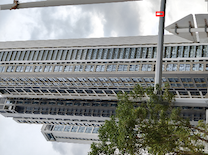}}\hspace{1pt}
\end{minipage}
\begin{minipage}[ht]{.99\linewidth}
\centering
\vspace{-12pt}
\subfloat[Input \textbf{W}]{\label{}\includegraphics[width=0.23\linewidth]{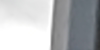}}\hspace{1pt}
\subfloat[Input \textbf{T}]{\label{}\includegraphics[width=0.23\linewidth]{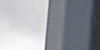}}\hspace{1pt}
\subfloat[W/o avg.]{\label{}\includegraphics[width=0.23\linewidth]{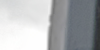}}\hspace{1pt}
\subfloat[With avg.]{\label{}\includegraphics[width=0.23\linewidth]{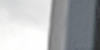}}\hspace{1pt}
\end{minipage}

\vspace{-10pt}

\caption{An example without and with the average of multiple forward warps of the {\bf{W}} images. This shows our choice can reduce artifacts caused by quantification errors of forward warp.}
\label{fig:multi_warps}
\end{figure}

\subsection{Transformation of {\bf{W}}}
The transformation is similar with the forward warp in Eq. \ref{eqn:Wfwd}. According to the flow changes between ${\bf{F}}_{bwd}^{{\bf{T}} \to {\bf{W}}}$ and ${\bf{\hat F}}_{bwd}^{{\bf{T}} \to {\bf{W}}}$, we forward warp 
${\bf{I}}^{\bf{W}}$ to the {\bf{O}} view result ${\bf{I}}_{\bf{W}}^{\bf{O}}$. The empty regions in the warped image ${\bf{I}}_{\bf{W}}^{\bf{O}}$ is not processed because they locate in non-occluded regions and are not used in the final blending. To mitigate artifacts, e.g. discontinuous lines, caused by quantization errors of forward warp, we introduce a subtle offset $(u, v)$ to the optical flow values. We perform multiple forward warps of ${\bf{I}}^{\bf{W}}$ using varying offset values and subsequently average the corresponding warping results to obtain the final outcome, i.e.
\begin{small}
\begin{equation}
{\bf{I}}_{\bf{W}}^{\bf{O}} = \frac{{\sum\limits_{u =  - 0.5}^{0.5} {\sum\limits_{v =  - 0.5}^{0.5} {War{p_{fwd}}({{\bf{I}}^{\bf{W}}},{\bf{\hat F}}_{bwd}^{{\bf{T}} \to {\bf{W}}} - {\bf{F}}_{bwd}^{{\bf{T}} \to {\bf{W}}} + (u,v))} } }}{{\sum\limits_{u =  - 0.5}^{0.5} {\sum\limits_{v =  - 0.5}^{0.5} 1 } }}
\end{equation}
\end{small}
where $u$ and $v$ change from -0.5 to 0.5 with the step of 0.2. As shown in Fig. \ref{fig:multi_warps}, the average of multiple warps can generate more natural edges but may blur the details. Since the area of occluded regions that need ${\bf{I}}_{\bf{W}}^{\bf{O}}$ is small, the blur will not affect the output a lot.

\subsection{Image Blending}
Details of occlusion computation, tone matching, and blending are introduced below.
\begin{figure}
\centering
\begin{minipage}[ht]{.99\linewidth}
\centering
\vspace{-0pt}
\subfloat[]{\label{}\includegraphics[width=0.18\linewidth]{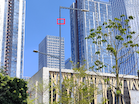}}\hspace{0.5pt}
\subfloat[]{\label{}\includegraphics[width=0.18\linewidth]{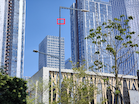}}\hspace{0.5pt}
\subfloat[]{\label{}\includegraphics[width=0.18\linewidth]{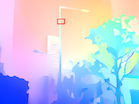}}\hspace{0.5pt}
\subfloat[]{\label{}\includegraphics[width=0.18\linewidth]{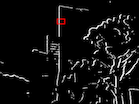}}\hspace{0.5pt}
\subfloat[]{\label{}\includegraphics[width=0.18\linewidth]{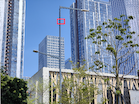}}\hspace{0.5pt}
\end{minipage}
\vspace{-0pt}
\begin{minipage}[ht]{.99\linewidth}
\centering
\vspace{-12pt}
\subfloat[Input \textbf{W}]{\label{}\includegraphics[width=0.18\linewidth]{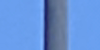}}\hspace{0.5pt}
\subfloat[Input \textbf{T}]{\label{}\includegraphics[width=0.18\linewidth]{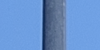}}\hspace{0.5pt}
\subfloat[Original]{\label{}\includegraphics[width=0.18\linewidth]{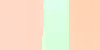}}\hspace{0.5pt}
\subfloat[Occlusion]
{\label{}\includegraphics[width=0.18\linewidth]{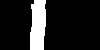}}\hspace{0.5pt}
\subfloat[Warped \textbf{T}]{\label{}\includegraphics[width=0.18\linewidth]{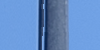}}\hspace{0.5pt}
\end{minipage}
 \footnotesize flow
\vspace{-10pt}

\caption{An example to show that the occlusion computed based on the optical flow can provide correct occlusion.}

\label{fig:occ}
\end{figure}

\textbf{Occlusion computation.} 
A baseline method for occlusion computation is the forward-backward flow consistency check \cite{lai2022face}, but it requires two times computation of optical flow. To reduce the computation of optical flow to only one time, i.e. only computing ${\bf{F}}_{bwd}^{{\bf{T}} \to {\bf{W}}}$, we propose the occlusion computation method based on a given backward optical flow map ${\bf{F}}_{bwd}$, i.e.
\begin{equation}
    {\bf{M}}_{occ} = OC({\bf{F}}_{bwd}),
\end{equation}
and ${\bf{M}}_{occ}^{\bf{W}}=OC({\bf{F}}_{bwd}^{{\bf{T}} \to {\bf{W}}})$ and ${\bf{M}}_{occ}^{\bf{F}}=OC({\bf{F}}_{bwd}^{{\bf{T}} \to {\bf{O}}})$. Due to calibration errors and slight relative movements of dual cameras during practical usage, the two cameras are not precisely aligned on the same baseline. In this paper, we assume the {\bf{T}} camera is positioned in the upper-left view, and the {\bf{W}} camera is in the lower-right view.
Based on the geometric relationship, objects in the foreground will generate occlusions on their left and upper sides. The area of occluded pixels is determined by the difference between the foreground and background optical flows. Therefore, we judge the pixels as foreground or background based on the magnitude of their optical flow values. For a given foreground pixel $(j, i)$, if its 8 neighboring pixels contain a background pixel $(y, x)$ in its left or upper directions,
the rectangular region from $(j-|{\bf{F}}_{bwd}(y,x,1)-{\bf{F}}_{bwd}(j,i,1)|,i-|{\bf{F}}_{bwd}(y,x,2)-{\bf{F}}_{bwd}(j,i,2)|)$ to $(j, i)$ is labeled as the occluded region. An example in Fig. \ref{fig:occ} shows the effectiveness of our occlusion computation method.

\captionsetup[subfloat]{labelsep=none,format=plain,labelformat=empty}
\begin{figure}
\vspace{-0pt}
\centering
\begin{minipage}[ht]{.99\linewidth}
\centering
\subfloat[]{\label{}\includegraphics[width=0.31\linewidth]{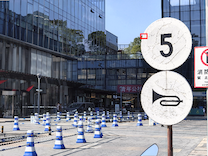}}\hspace{1pt}
\subfloat[]{\label{}\includegraphics[width=0.31\linewidth]{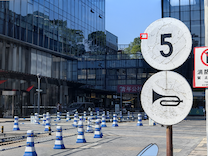}}\hspace{1pt}
\subfloat[]{\label{}\includegraphics[width=0.31\linewidth]{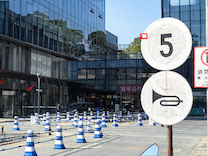}}\hspace{1pt}
\end{minipage}
\begin{minipage}[ht]{.99\linewidth}
\centering
\vspace{-12pt}
\subfloat[Global]{\label{}\includegraphics[width=0.31\linewidth]{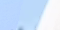}}\hspace{1pt}
\subfloat[Poisson]{\label{}\includegraphics[width=0.31\linewidth]{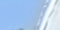}}\hspace{1pt}
\subfloat[Ours]{\label{}\includegraphics[width=0.31\linewidth]{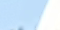}}\hspace{1pt}
\end{minipage}
\vspace{-10pt}
\caption{An example to show regional histogram equalization is better than global adjustment and Poisson blending in tone matching.}

\label{fig:color}

\end{figure}

\textbf{Tone matching.} 
We adopt regional histogram equalization to adjust the tone of ${\bf{I}}_{\bf{T}}^{\bf{O}}$ to match the tone of ${\bf{I}}_{\bf{W}}^{\bf{O}}$, i.e.
\begin{equation}
    {\bf{I}}_{\bf{T}}^{{\bf{O'}}} = RHE({\bf{I}}_{\bf{T}}^{\bf{O}},{\bf{I}}_{\bf{W}}^{\bf{O}}),
\end{equation}
where we segment ${\bf{I}}_{\bf{T}}^{\bf{O}}$ and ${\bf{I}}_{\bf{W}}^{\bf{O}}$ into overlapped blocks with the block size of $200 \times 200$ and the stride of $30$. In each block, we perform histogram equalization \cite{histogram} of the {\bf{T}} block to transform its intensity so that the histogram of the output block approximately matches the histogram of the {\bf{W}} block.

\textbf{Blending.} 
We blend ${\bf{I}}_{\bf{T}}^{{\bf{O'}}}$ in non-occluded regions with ${\bf{I}}_{\bf{W}}^{{\bf{O}}}$ in occluded regions to obtain the final output image ${{\bf{I}}^{\bf{O}}}$ via pyramid blending \cite{mertens2007exposure}, i.e.
\begin{equation}
{{\bf{I}}^{\bf{O}}} = PyramidBlending({\bf{I}}_{\bf{W}}^{\bf{O}},{\bf{I}}_{\bf{T}}^{{\bf{O'}}},{\bf{M}}_{occ}^{\bf{O}}),
\end{equation}
where we set the blending weight value as 1 in the occluded region and set a 15-pixels-width soft-blending zone neighboring the occluded region to let the weight value change linearly from 1 to 0. The weight value of the rest region is set as 0. The full-view output ${{\bf{I}}^{\bf{F}}}$ is obtained by blending the estimated overlap-region result ${\bf{I}}^{\bf{O}}$ with the original wide image ${\bf{I}}_{\bf{W}}^{{\bf{F}}}$ outside the overlap-region, and the mask map ${\bf{M}}_{ol}^{\bf{F}}$ masks the overlap region, i.e.
\begin{equation}
{{\bf{I}}^{\bf{F}}} = PyramidBlending({\bf{I}}^{\bf{O}},I_{\bf{W}}^{{\bf{F}}},{\bf{M}}_{ol}^{\bf{F}}),
\end{equation}
where we set a 100-pixels-width soft-blending weights in the boundary zone of the overlap region in ${\bf{M}}_{ol}^{\bf{F}}$.
\section{Results}
\subsection{Datasets}

We use two datasets, i.e. CameraFusion \cite{dcsr} and OPPO72. CameraFusion is a public dual camera image fusion dataset with 131 train pairs and 15 test pairs. OPPO72 is a dataset with 72 pairs of simultaneously shot {\bf{W}} and {\bf{T}} images that we collect using the {\bf{W}} and {\bf{T}} cameras of an OPPO Find X6 cellphone.

For training the comparison algorithms, we use the training set of CameraFusion. For testing different algorithms, we use the testing set of CameraFusion and the OPPO72 dataset. We do not train the comparison algorithms on the OPPO72 dataset, aiming to examine their generalization capabilities. 

Due to the geometric transformations of image contents introduced in our method, conventional full-reference metrics are not suitable for quality evaluation. Therefore, we use no-reference metrics of Brisque \cite{mittal2011blind}, NIQE \cite{mittal2012making}, NRQM \cite{ma2017learning}, and PI \cite{ignatov2018pirm}, for objective quality assessment.

\begin{table}
    \centering
    \setlength{\tabcolsep}{1pt}
    \renewcommand\arraystretch{0.6}
    \begin{tabular}{c|cccc|cccc}
    \toprule
    & \multicolumn{4}{c|}{OPPO72}&\multicolumn{4}{c}{CameraFusion} \\
        \midrule
         Method&  Brisque&  NIQE&  NRQM& PI&  Brisque&  NIQE& NRQM& PI\\
         \midrule
         EDSR& 57.36 & 6.23 & 4.76 & 5.81&  56.45 & 5.94 & 4.71 & 5.69\\
         RCAN& 57.03 & 6.33 & 4.71 & 5.87& 56.68 & 5.99 & 4.74 & 5.72\\
         SwinIR& 54.31 & 4.96 & 5.79 & 4.61& 57.85 & 5.21 & 5.77 & 4.81\\
         Real-E& 28.13 & 4.39 & 6.07 & 4.26& 30.33 & 4.46 & 5.97 & 4.37
         \\
          \toprule
         SRNTT& 42.52 & 5.80 & 4.98 & 5.45& \textbf{11.89} & 5.94 & 6.05 & 4.99\\
         TTSR& 29.96 & 4.62 & 5.75 & 4.52 & 12.46 & 3.96 & 6.35 & 3.89\\
         MASA& 23.70 & 4.17 & 5.98 & 4.21 & 38.32 & 5.62 & 5.44 & 5.19\\
         C$^{2}$& 34.39 & 4.70 & 5.45 & 4.72& 30.82 & 4.38 & 5.83 & 4.46\\
         \toprule
         DCSR& 31.47 & 5.83 & 5.59 & 5.32& 29.75 & 6.37 & 6.35 & 5.35\\
 DZSR& 35.51 & 5.83 & 5.72 & 5.12& 55.73 & 6.34 & 5.06 & 5.80\\
 ZeDuSR& 39.14 & 5.65 & 5.26 & 5.31& 35.24 & 5.32 & 5.91 & 4.85\\
 \toprule
 Ours& \textbf{17.87} & \textbf{4.20} & \textbf{6.08} & \textbf{4.18}& 20.48 & \textbf{3.72} & \textbf{6.37} & \textbf{3.81}\\
 \bottomrule
    \end{tabular}
    \vspace{-8pt}
    \caption{Quantitative results. C$^{2}$ is short for C$^{2}$-Matching, Real-E is short for Real-ESRGAN, and DZSR is short for SelfDZSR.}
    \label{tab:quantitative_result}
\end{table}

\captionsetup[subfloat]{labelsep=none,format=plain,labelformat=empty}
\begin{figure}
\vspace{-0pt}
\centering
\begin{minipage}[ht]{.99\linewidth}
\centering
\subfloat[Clarity score of OPPO72]{\label{}\includegraphics[width=0.48\linewidth]{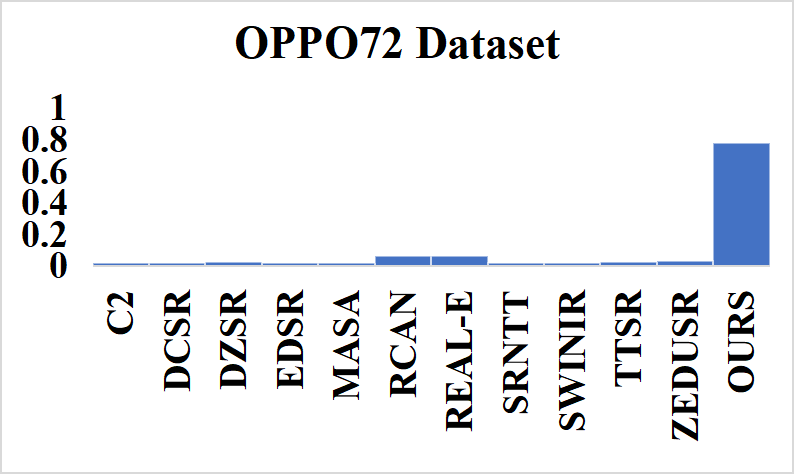}}\hspace{4pt}
\subfloat[Clarity score of CameraFusion]{\label{}\includegraphics[width=0.48\linewidth]{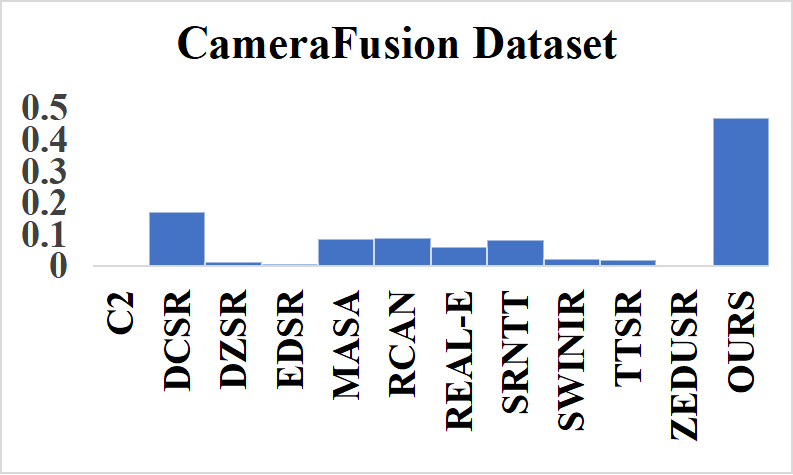}}\hspace{4pt}
\end{minipage}
\begin{minipage}[ht]{.99\linewidth}
\centering
\subfloat[Artifacts score of OPPO72]{\label{}\includegraphics[width=0.48\linewidth]{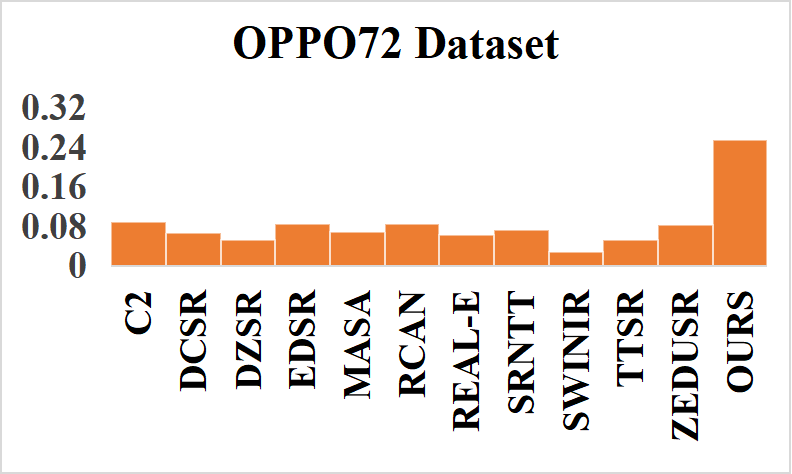}}\hspace{4pt}
\subfloat[Artifacts score of CameraFusion]{\label{}\includegraphics[width=0.48\linewidth]{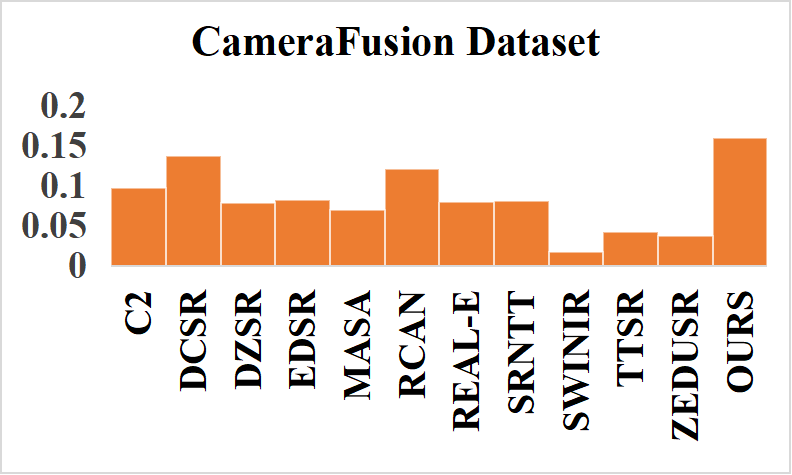}}\hspace{4pt}
\end{minipage}
\vspace{-10pt}
\caption{User study. C2 is short for C$^{2}$-Matching, Real-E is short for Real-ESRGAN, and DZSR is short for SelfDZSR.}
\label{fig:user_study}
\end{figure}

\subsection{Comparison Algorithms}
We compared 11 state-of-the-art algorithms, including single image super-resolution algorithms such as EDSR \cite{EDSR}, RCAN \cite{RCAN}, SwinIR \cite{SwinIR}, and Real-ESRGAN \cite{realesrgan}, reference-based super-resolution algorithms such as SRNTT \cite{SRNTT}, TTSR \cite{TTSR}, MASA \cite{masa} and C$^{2}$-Matching \cite{c2_matching}, and dual-camera super-resolution algorithms such as DCSR \cite{dcsr}, SelfDZSR \cite{dzsr} and ZeDuSR \cite{zedusr}.

\subsection{Results and Result Analysis}
Quantitative and qualitative results are presented in Table \ref{tab:quantitative_result} and Figs. \ref{fig:OPPO72_Res} \ref{fig:CameraFusion_res}. From the results, it can be observed that the output of comparison algorithms is at the {\bf{W}} view. These methods are deep learning-based and typically employ loss functions aimed at improving the average quality. However, due to the limitations of the optimization process and training data, the trained networks exhibit inferior performance not only in occluded areas but also in non-occluded regions, compared to the quality of the {\bf{T}} image.

In contrast, our algorithm directly uses the pixels of the warped {\bf{T}} image into the non-occluded regions of the output and minimizes occluded regions by leveraging view transition. The occlusion areas on the OPPO72 and CameraFusion are 12.71\% and 17.75\% respectively without our view transition and 4.27\% and 4.28\% respectively with our view transition. This results in the increased usage of {\bf{T}} pixels into the output and thus significant quality advantage in comparison with the comparison methods.

\captionsetup[subfloat]{labelsep=none,format=plain,labelformat=empty}
\begin{figure*}
\begin{minipage}[ht]{.99\linewidth}
\centering
\subfloat{\label{}\includegraphics[width=0.18\linewidth, height= 0.1\linewidth]{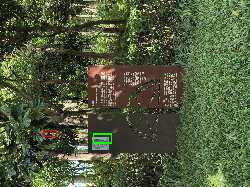}}\hspace{4pt}
\subfloat{\label{}\includegraphics[width=0.18\linewidth, height= 0.1\linewidth]{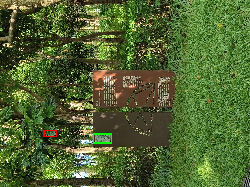}}\hspace{4pt}
\subfloat{\label{}\includegraphics[width=0.18\linewidth, height= 0.1\linewidth]{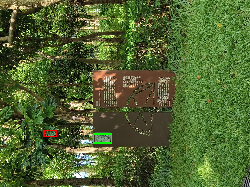}}\hspace{4pt}
\subfloat{\label{}\includegraphics[width=0.18\linewidth, height= 0.1\linewidth]{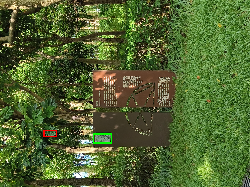}}\hspace{4pt}
\subfloat{\label{}\includegraphics[width=0.18\linewidth, height= 0.1\linewidth]{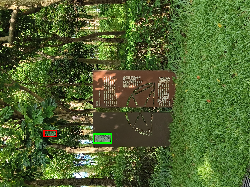}}\hspace{4pt}
\end{minipage}
\begin{minipage}[ht]{.99\linewidth}
\centering
\subfloat{\label{}\includegraphics[width=0.18\linewidth]{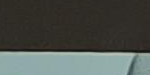}}\hspace{4pt}
\subfloat{\label{}\includegraphics[width=0.18\linewidth]{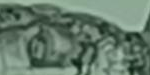}}\hspace{4pt}
\subfloat{\label{}\includegraphics[width=0.18\linewidth]{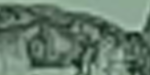}}\hspace{4pt}
\subfloat{\label{}\includegraphics[width=0.18\linewidth]{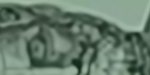}}\hspace{4pt}
\subfloat{\label{}\includegraphics[width=0.18\linewidth]{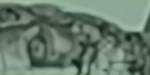}}\hspace{4pt}
\end{minipage}
\begin{minipage}[ht]{.99\linewidth}
\centering
\subfloat[Input \textbf{T}]{\label{}\includegraphics[width=0.18\linewidth]{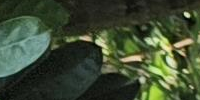}}\hspace{4pt}
\subfloat[Input \textbf{W}]{\label{}\includegraphics[width=0.18\linewidth]{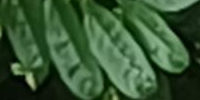}}\hspace{4pt}
\subfloat[Input \textbf{W}-4x]{\label{}\includegraphics[width=0.18\linewidth]{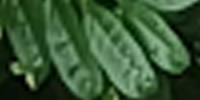}}\hspace{4pt}
\subfloat[EDSR]{\label{}\includegraphics[width=0.18\linewidth]{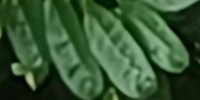}}\hspace{4pt}
\subfloat[RCAN]{\label{}\includegraphics[width=0.18\linewidth]{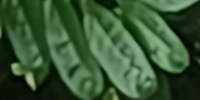}}\hspace{4pt}
\end{minipage}
\begin{minipage}[ht]{.99\linewidth}
\centering
\subfloat{\label{}\includegraphics[width=0.18\linewidth, height= 0.1\linewidth]{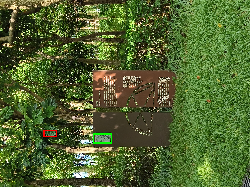}}\hspace{4pt}
\subfloat{\label{}\includegraphics[width=0.18\linewidth, height= 0.1\linewidth]{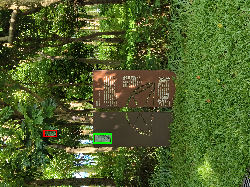}}\hspace{4pt}
\subfloat{\label{}\includegraphics[width=0.18\linewidth, height= 0.1\linewidth]{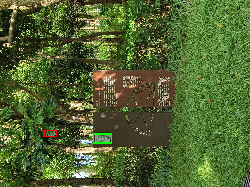}}\hspace{4pt}
\subfloat{\label{}\includegraphics[width=0.18\linewidth, height= 0.1\linewidth]{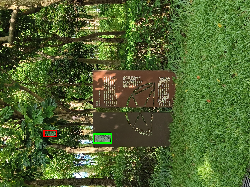}}\hspace{4pt}
\subfloat{\label{}\includegraphics[width=0.18\linewidth, height= 0.1\linewidth]{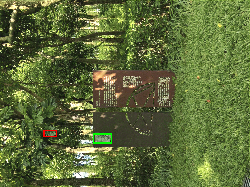}}\hspace{4pt}
\end{minipage}
\begin{minipage}[ht]{.99\linewidth}
\centering
\subfloat{\label{}\includegraphics[width=0.18\linewidth]{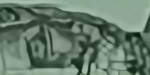}}\hspace{4pt}
\subfloat{\label{}\includegraphics[width=0.18\linewidth]{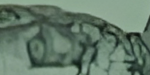}}\hspace{4pt}
\subfloat{\label{}\includegraphics[width=0.18\linewidth]{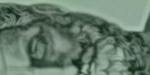}}\hspace{4pt}
\subfloat{\label{}\includegraphics[width=0.18\linewidth]{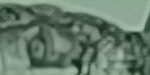}}\hspace{4pt}
\subfloat{\label{}\includegraphics[width=0.18\linewidth]{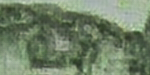}}\hspace{4pt}
\end{minipage}
\end{figure*}
\captionsetup[subfloat]{labelsep=none,format=plain,labelformat=empty}
\begin{figure*}
\setcounter{figure}{13}
\vspace{-11pt}
\begin{minipage}[ht]{.99\linewidth}
\centering
\subfloat[SwinIR]{\label{}\includegraphics[width=0.18\linewidth]{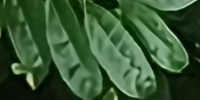}}\hspace{4pt}
\subfloat[Real-ESRGAN]{\label{}\includegraphics[width=0.18\linewidth]{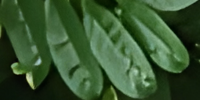}}\hspace{4pt}
\subfloat[C$^{2}$-Matching]{\label{}\includegraphics[width=0.18\linewidth]{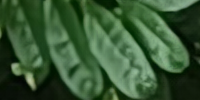}}\hspace{4pt}
\subfloat[SRNTT]{\label{}\includegraphics[width=0.18\linewidth]{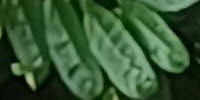}}\hspace{4pt}
\subfloat[TTSR]{\label{}\includegraphics[width=0.18\linewidth]{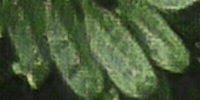}}\hspace{4pt}
\end{minipage}
\begin{minipage}[h]{.99\linewidth}
\centering
\subfloat{\label{}\includegraphics[width=0.18\linewidth, height= 0.1\linewidth]{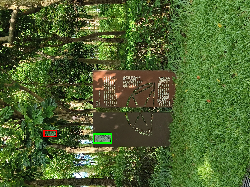}}\hspace{4pt}
\subfloat{\label{}\includegraphics[width=0.18\linewidth, height= 0.1\linewidth]{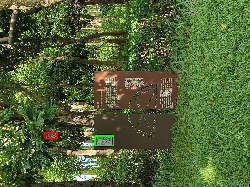}}\hspace{4pt}
\subfloat{\label{}\includegraphics[width=0.18\linewidth, height= 0.1\linewidth]{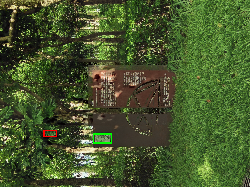}}\hspace{4pt}
\subfloat{\label{}\includegraphics[width=0.18\linewidth, height= 0.1\linewidth]{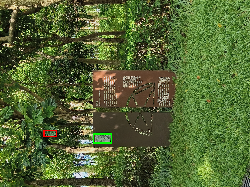}}\hspace{4pt}
\subfloat{\label{}\includegraphics[width=0.18\linewidth, height= 0.1\linewidth]{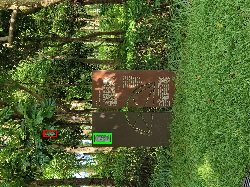}}\hspace{4pt}
\end{minipage}

\begin{minipage}[ht]{.99\linewidth}
\centering
\subfloat{\label{}\includegraphics[width=0.18\linewidth]{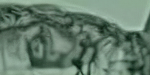}}\hspace{4pt}
\subfloat{\label{}\includegraphics[width=0.18\linewidth]{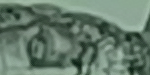}}\hspace{4pt}
\subfloat{\label{}\includegraphics[width=0.18\linewidth]{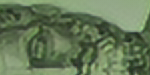}}\hspace{4pt}
\subfloat{\label{}\includegraphics[width=0.18\linewidth]{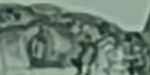}}\hspace{4pt}
\subfloat{\label{}\includegraphics[width=0.18\linewidth]{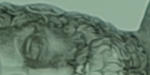}}\hspace{4pt}
\end{minipage}
\begin{minipage}[ht]{.99\linewidth}
\centering
\subfloat[MASA]{\label{}\includegraphics[width=0.18\linewidth]{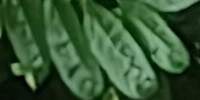}}\hspace{4pt}
\subfloat[DCSR]{\label{}\includegraphics[width=0.18\linewidth]{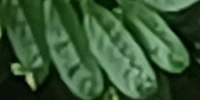}}\hspace{4pt}
\subfloat[SelfDZSR]{\label{}\includegraphics[width=0.18\linewidth]{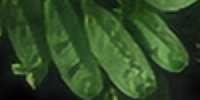}}\hspace{4pt}
\subfloat[ZeDuSR]{\label{}\includegraphics[width=0.18\linewidth]{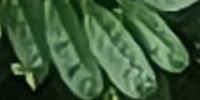}}\hspace{4pt}
\subfloat[Ours]{\label{}\includegraphics[width=0.18\linewidth]{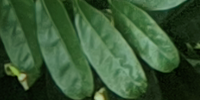}}\hspace{4pt}
\end{minipage}
\caption{Example results of different methods on the OPPO72 dataset. Regions of red and green boxes are enlarged.}
\label{fig:OPPO72_Res}
\end{figure*}
\subsection{Ablation}
We provide intermediate results in Figs. \ref{fig:Wfwd_vs_Wbwd} \ref{fig:box_filter} \ref{fig:distance} \ref{fig:ratio} \ref{fig:empty} \ref{fig:multi_warps} \ref{fig:occ} \ref{fig:color}, where we conduct processing and parameter variations on critical steps in our pipeline to demonstrate the rationality of our choices.

\captionsetup[subfloat]{labelsep=none,format=plain,labelformat=empty}
\begin{figure*}
\vspace{-10pt}
\begin{minipage}[t]{.99\linewidth}
\centering
\subfloat{\label{}\includegraphics[width=0.18\linewidth, height= 0.1\linewidth]{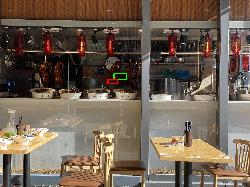}}\hspace{4pt}
\subfloat{\label{}\includegraphics[width=0.18\linewidth, height= 0.1\linewidth]{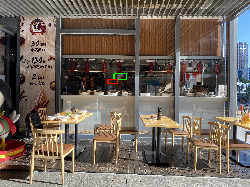}}\hspace{4pt}
\subfloat{\label{}\includegraphics[width=0.18\linewidth, height= 0.1\linewidth]{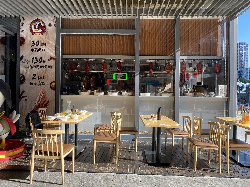}}\hspace{4pt}
\subfloat{\label{}\includegraphics[width=0.18\linewidth, height= 0.1\linewidth]{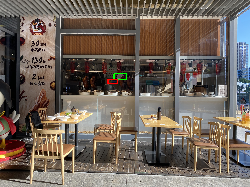}}\hspace{4pt}
\subfloat{\label{}\includegraphics[width=0.18\linewidth, height= 0.1\linewidth]{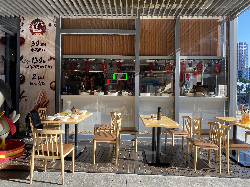}}\hspace{4pt}
\end{minipage}
\begin{minipage}[ht]{.99\linewidth}
\centering
\subfloat{\label{}\includegraphics[width=0.18\linewidth, height= 0.1\linewidth]{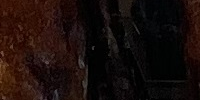}}\hspace{4pt}
\subfloat{\label{}\includegraphics[width=0.18\linewidth, height= 0.1\linewidth]{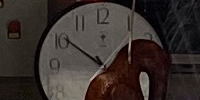}}\hspace{4pt}
\subfloat{\label{}\includegraphics[width=0.18\linewidth, height= 0.1\linewidth]{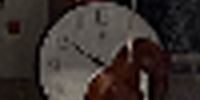}}\hspace{4pt}
\subfloat{\label{}\includegraphics[width=0.18\linewidth, height= 0.1\linewidth]{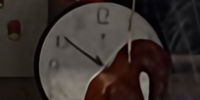}}\hspace{4pt}
\subfloat{\label{}\includegraphics[width=0.18\linewidth, height= 0.1\linewidth]{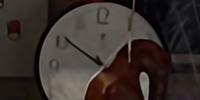}}\hspace{4pt}
\end{minipage}
\begin{minipage}[ht]{.99\linewidth}
\centering
\subfloat[Input \textbf{T}]{\label{}\includegraphics[width=0.18\linewidth]{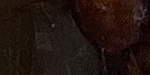}}\hspace{4pt}
\subfloat[Input \textbf{W}]{\label{}\includegraphics[width=0.18\linewidth]{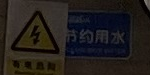}}\hspace{4pt}
\subfloat[Input \textbf{W}-4x]{\label{}\includegraphics[width=0.18\linewidth]{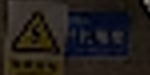}}\hspace{4pt}
\subfloat[EDSR]{\label{}\includegraphics[width=0.18\linewidth]{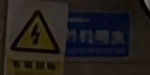}}\hspace{4pt}
\subfloat[RCAN]{\label{}\includegraphics[width=0.18\linewidth]{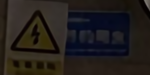}}\hspace{4pt}
\end{minipage}
\begin{minipage}[ht]{.99\linewidth}
\centering
\subfloat{\label{}\includegraphics[width=0.18\linewidth, height= 0.1\linewidth]{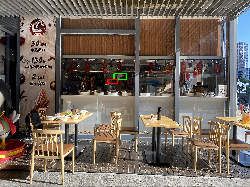}}\hspace{4pt}
\subfloat{\label{}\includegraphics[width=0.18\linewidth, height= 0.1\linewidth]{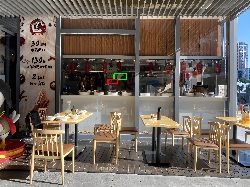}}\hspace{4pt}
\subfloat{\label{}\includegraphics[width=0.18\linewidth, height= 0.1\linewidth]{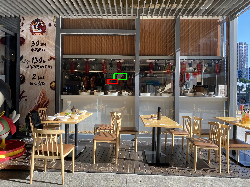}}\hspace{4pt}
\subfloat{\label{}\includegraphics[width=0.18\linewidth, height= 0.1\linewidth]{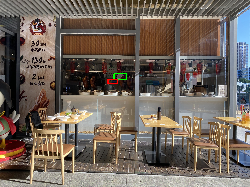}}\hspace{4pt}
\subfloat{\label{}\includegraphics[width=0.18\linewidth, height= 0.1\linewidth]{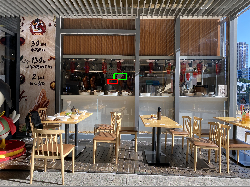}}\hspace{4pt}
\end{minipage}
\begin{minipage}[ht]{.99\linewidth}
\centering
\subfloat{\label{}\includegraphics[width=0.18\linewidth, height= 0.1\linewidth]{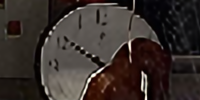}}\hspace{4pt}
\subfloat{\label{}\includegraphics[width=0.18\linewidth, height= 0.1\linewidth]{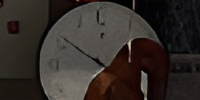}}\hspace{4pt}
\subfloat{\label{}\includegraphics[width=0.18\linewidth, height= 0.1\linewidth]{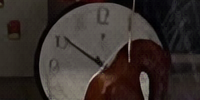}}\hspace{4pt}
\subfloat{\label{}\includegraphics[width=0.18\linewidth, height= 0.1\linewidth]{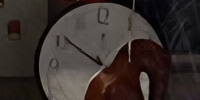}}\hspace{4pt}
\subfloat{\label{}\includegraphics[width=0.18\linewidth, height= 0.1\linewidth]{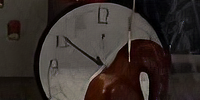}}\hspace{4pt}
\end{minipage}
\end{figure*}

\captionsetup[subfloat]{labelsep=none,format=plain,labelformat=empty}
\begin{figure*}
\vspace{-11pt}
\setcounter{figure}{14}
\begin{minipage}[ht]{.99\linewidth}
\centering
\subfloat[SwinIR]{\label{}\includegraphics[width=0.18\linewidth]{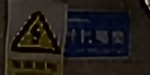}}\hspace{4pt}
\subfloat[Real-ESRGAN]{\label{}\includegraphics[width=0.18\linewidth]{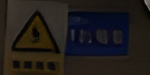}}\hspace{4pt}
\subfloat[C$^{2}$-Matching]{\label{}\includegraphics[width=0.18\linewidth]{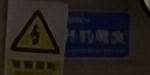}}\hspace{4pt}
\subfloat[SRNTT]{\label{}\includegraphics[width=0.18\linewidth]{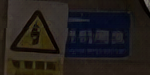}}\hspace{4pt}
\subfloat[TTSR]{\label{}\includegraphics[width=0.18\linewidth]{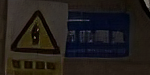}}\hspace{4pt}
\end{minipage}
\begin{minipage}[ht]{.99\linewidth}
\centering
\subfloat{\label{}\includegraphics[width=0.18\linewidth, height= 0.1\linewidth]{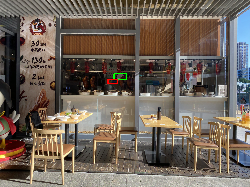}}\hspace{4pt}
\subfloat{\label{}\includegraphics[width=0.18\linewidth, height= 0.1\linewidth]{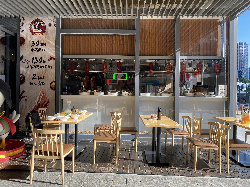}}\hspace{4pt}
\subfloat{\label{}\includegraphics[width=0.18\linewidth, height= 0.1\linewidth]{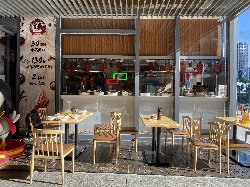}}\hspace{4pt}
\subfloat{\label{}\includegraphics[width=0.18\linewidth, height= 0.1\linewidth]{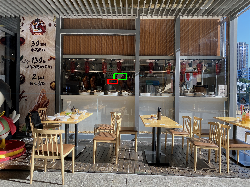}}\hspace{4pt}
\subfloat{\label{}\includegraphics[width=0.18\linewidth, height= 0.1\linewidth]{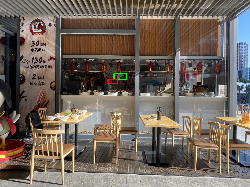}}\hspace{4pt}
\end{minipage}
\begin{minipage}[ht]{.99\linewidth}
\centering
\subfloat{\label{}\includegraphics[width=0.18\linewidth, height= 0.1\linewidth]{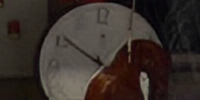}}\hspace{4pt}
\subfloat{\label{}\includegraphics[width=0.18\linewidth, height= 0.1\linewidth]{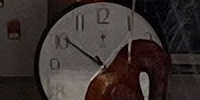}}\hspace{4pt}
\subfloat{\label{}\includegraphics[width=0.18\linewidth, height= 0.1\linewidth]{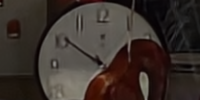}}\hspace{4pt}
\subfloat{\label{}\includegraphics[width=0.18\linewidth, height= 0.1\linewidth]{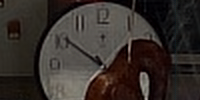}}\hspace{4pt}
\subfloat{\label{}\includegraphics[width=0.18\linewidth, height= 0.1\linewidth]{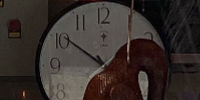}}\hspace{4pt}
\end{minipage}
\begin{minipage}[ht]{.99\linewidth}
\centering
\subfloat[MASA]{\label{}\includegraphics[width=0.18\linewidth]{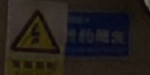}}\hspace{4pt}
\subfloat[DCSR]{\label{}\includegraphics[width=0.18\linewidth]{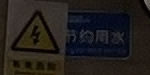}}\hspace{4pt}
\subfloat[SelfDZSR]{\label{}\includegraphics[width=0.18\linewidth]{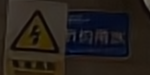}}\hspace{4pt}
\subfloat[ZeDuSR]{\label{}\includegraphics[width=0.18\linewidth]{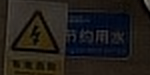}}\hspace{4pt}
\subfloat[Ours]{\label{}\includegraphics[width=0.18\linewidth]{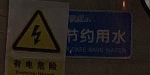}}\hspace{4pt}
\end{minipage}
\caption{Example results of different methods on the CameraFusion dataset. Regions of red and green boxes are enlarged.}
\label{fig:CameraFusion_res}
\end{figure*}

\subsection{User Study}
As the primary users of mobile imaging are humans, we conduct a user study with 30 participants to evaluate the enhancement quality of different algorithms. Our user study considers two aspects of image quality: image clarity and artifacts (e.g. inconsistent color tones, ghosting, unnatural textures). During the user study, to ensure that the displayed images have sufficient clarity in details and are easily observable, we present cropped patches of $250 \times 250$ pixels from the images and simultaneously displayed the results of 12 algorithms on one screen. 6 images of the CameraFusion testing set and 20 images of the OPPO72 dataset are randomly selected, with 10 randomly cropped patches from each image, resulting in a total of 260 sets of cropped patches. Users are asked to compare these results and select one or more that exhibited the best performance. The average scores are shown in Figure \ref{fig:user_study}, and our algorithm achieves higher scores than other SOTA methods in terms of both image clarity and artifacts.

\section{Limitations and Conclusions}
Limitations: 1) Our method can only enhance the quality of the overlap region of the {\bf{W}} and {\bf{T}} images. The non-overlap region is not enhanced at all by our method and it can be enhanced by the other super-resolution methods before blending with our output of the overlap region. 2) Our method relies on accurate optical flow from the {\bf{T}} image to the {\bf{W}} image. When the estimated optical flow is not correct, our processing will generate artifacts in the output.

Conclusions: We propose a view transition method to fuse the {\bf{W}} and {\bf{T}} images from the dual camera system of smart phones to generate higher quality output in their overlap region. We position the output image in a mixed view of {\bf{W}} and {\bf{T}} images to maximize the utilization of high-quality pixels from the {\bf{T}} image and ensure no obvious distortion of content shapes. Experimental results show the superiority quality of our method than the SOTA methods.
\clearpage

\bibliographystyle{IEEEtran}
\bibliography{main}

\begin{thebibliography}{10}
\providecommand{\url}[1]{#1}
\csname url@samestyle\endcsname
\providecommand{\newblock}{\relax}
\providecommand{\bibinfo}[2]{#2}
\providecommand{\BIBentrySTDinterwordspacing}{\spaceskip=0pt\relax}
\providecommand{\BIBentryALTinterwordstretchfactor}{4}
\providecommand{\BIBentryALTinterwordspacing}{\spaceskip=\fontdimen2\font plus
\BIBentryALTinterwordstretchfactor\fontdimen3\font minus \fontdimen4\font\relax}
\providecommand{\BIBforeignlanguage}[2]{{%
\expandafter\ifx\csname l@#1\endcsname\relax
\typeout{** WARNING: IEEEtran.bst: No hyphenation pattern has been}%
\typeout{** loaded for the language `#1'. Using the pattern for}%
\typeout{** the default language instead.}%
\else
\language=\csname l@#1\endcsname
\fi
#2}}
\providecommand{\BIBdecl}{\relax}
\BIBdecl

\bibitem{lai2022face}
W.-S. Lai, Y.~Shih, L.-C. Chu, X.~Wu, S.-F. Tsai, M.~Krainin, D.~Sun, and C.-K. Liang, ``Face deblurring using dual camera fusion on mobile phones,'' \emph{ACM Transactions on Graphics (TOG)}, vol.~41, no.~4, pp. 1--16, 2022.

\bibitem{dcsr}
T.~Wang, J.~Xie, W.~Sun, Q.~Yan, and Q.~Chen, ``Dual-camera super-resolution with aligned attention modules,'' in \emph{Proceedings of the IEEE/CVF International Conference on Computer Vision}, 2021, pp. 2001--2010.

\bibitem{histogram}
S.~M. Pizer, E.~P. Amburn, J.~D. Austin, R.~Cromartie, A.~Geselowitz, T.~Greer, B.~ter Haar~Romeny, J.~B. Zimmerman, and K.~Zuiderveld, ``Adaptive histogram equalization and its variations,'' \emph{Computer vision, graphics, and image processing}, vol.~39, no.~3, pp. 355--368, 1987.

\bibitem{mertens2007exposure}
T.~Mertens, J.~Kautz, and F.~Van~Reeth, ``Exposure fusion,'' in \emph{15th Pacific Conference on Computer Graphics and Applications (PG'07)}.\hskip 1em plus 0.5em minus 0.4em\relax IEEE, 2007, pp. 382--390.

\bibitem{dzsr}
Z.~Zhang, R.~Wang, H.~Zhang, Y.~Chen, and W.~Zuo, ``Self-supervised learning for real-world super-resolution from dual zoomed observations,'' in \emph{European Conference on Computer Vision}.\hskip 1em plus 0.5em minus 0.4em\relax Springer, 2022, pp. 610--627.

\bibitem{zedusr}
R.~Xu, M.~Yao, and Z.~Xiong, ``Zero-shot dual-lens super-resolution,'' in \emph{Proceedings of the IEEE/CVF Conference on Computer Vision and Pattern Recognition}, 2023, pp. 9130--9139.

\bibitem{TTSR}
F.~Yang, H.~Yang, J.~Fu, H.~Lu, and B.~Guo, ``Learning texture transformer network for image super-resolution,'' in \emph{Proceedings of the IEEE/CVF conference on computer vision and pattern recognition}, 2020, pp. 5791--5800.

\bibitem{SRNTT}
Z.~Zhang, Z.~Wang, Z.~Lin, and H.~Qi, ``Image super-resolution by neural texture transfer,'' in \emph{Proceedings of the IEEE/CVF conference on computer vision and pattern recognition}, 2019, pp. 7982--7991.

\bibitem{resnet}
K.~He, X.~Zhang, S.~Ren, and J.~Sun, ``Deep residual learning for image recognition,'' in \emph{Proceedings of the IEEE conference on computer vision and pattern recognition}, 2016, pp. 770--778.

\bibitem{masa}
L.~Lu, W.~Li, X.~Tao, J.~Lu, and J.~Jia, ``Masa-sr: Matching acceleration and spatial adaptation for reference-based image super-resolution,'' in \emph{Proceedings of the IEEE/CVF Conference on Computer Vision and Pattern Recognition}, 2021, pp. 6368--6377.

\bibitem{c2_matching}
G.~Shim, J.~Park, and I.~S. Kweon, ``Robust reference-based super-resolution with similarity-aware deformable convolution,'' in \emph{Proceedings of the IEEE/CVF conference on computer vision and pattern recognition}, 2020, pp. 8425--8434.

\bibitem{dai2017deformable}
J.~Dai, H.~Qi, Y.~Xiong, Y.~Li, G.~Zhang, H.~Hu, and Y.~Wei, ``Deformable convolutional networks,'' in \emph{Proceedings of the IEEE international conference on computer vision}, 2017, pp. 764--773.

\bibitem{RCAN}
Y.~Zhang, K.~Li, K.~Li, L.~Wang, B.~Zhong, and Y.~Fu, ``Image super-resolution using very deep residual channel attention networks,'' in \emph{Proceedings of the European conference on computer vision (ECCV)}, 2018, pp. 286--301.

\bibitem{EDSR}
B.~Lim, S.~Son, H.~Kim, S.~Nah, and K.~Mu~Lee, ``Enhanced deep residual networks for single image super-resolution,'' in \emph{Proceedings of the IEEE conference on computer vision and pattern recognition workshops}, 2017, pp. 136--144.

\bibitem{SwinIR}
J.~Liang, J.~Cao, G.~Sun, K.~Zhang, L.~Van~Gool, and R.~Timofte, ``Swinir: Image restoration using swin transformer,'' in \emph{Proceedings of the IEEE/CVF international conference on computer vision}, 2021, pp. 1833--1844.

\bibitem{realesrgan}
X.~Wang, L.~Xie, C.~Dong, and Y.~Shan, ``Real-esrgan: Training real-world blind super-resolution with pure synthetic data,'' in \emph{Proceedings of the IEEE/CVF international conference on computer vision}, 2021, pp. 1905--1914.

\bibitem{srgan}
C.~Ledig, L.~Theis, F.~Husz{\'a}r, J.~Caballero, A.~Cunningham, A.~Acosta, A.~Aitken, A.~Tejani, J.~Totz, Z.~Wang \emph{et~al.}, ``Photo-realistic single image super-resolution using a generative adversarial network,'' in \emph{Proceedings of the IEEE conference on computer vision and pattern recognition}, 2017, pp. 4681--4690.

\bibitem{smith1992correction}
W.~E. Smith, N.~Vakil, and S.~A. Maislin, ``Correction of distortion in endoscope images,'' \emph{IEEE Transactions on Medical Imaging}, vol.~11, no.~1, pp. 117--122, 1992.

\bibitem{Reinhard}
E.~Reinhard, M.~Stark, P.~Shirley, and J.~Ferwerda, ``Photographic tone reproduction for digital images,'' \emph{ACM Transactions on Graphics}, vol.~21, 05 2002.

\bibitem{poisson}
P.~P{\'e}rez, M.~Gangnet, and A.~Blake, ``Poisson image editing,'' in \emph{Seminal Graphics Papers: Pushing the Boundaries, Volume 2}, 2023, pp. 577--582.

\bibitem{huang2022flowformer}
Z.~Huang, X.~Shi, C.~Zhang, Q.~Wang, K.~C. Cheung, H.~Qin, J.~Dai, and H.~Li, ``Flowformer: A transformer architecture for optical flow,'' in \emph{European Conference on Computer Vision}.\hskip 1em plus 0.5em minus 0.4em\relax Springer, 2022, pp. 668--685.

\bibitem{szeliski2022computer}
R.~Szeliski, \emph{Computer vision: algorithms and applications}.\hskip 1em plus 0.5em minus 0.4em\relax Springer Nature, 2022.

\bibitem{distortion}
``Radial distortion coefficient of arducam camera lens.'' \url{https://www.arducam.com/product/arducam-1-2-5-m12-mount-4mm-focal-length-low-distortion-camera-lens-m2504zh05s/}, (2023, Nov 16).

\bibitem{mittal2011blind}
A.~Mittal, A.~K. Moorthy, and A.~C. Bovik, ``Blind/referenceless image spatial quality evaluator,'' in \emph{2011 conference record of the forty fifth asilomar conference on signals, systems and computers (ASILOMAR)}.\hskip 1em plus 0.5em minus 0.4em\relax IEEE, 2011, pp. 723--727.

\bibitem{mittal2012making}
A.~Mittal, R.~Soundararajan, and A.~C. Bovik, ``Making a “completely blind” image quality analyzer,'' \emph{IEEE Signal processing letters}, vol.~20, no.~3, pp. 209--212, 2012.

\bibitem{ma2017learning}
C.~Ma, C.-Y. Yang, X.~Yang, and M.-H. Yang, ``Learning a no-reference quality metric for single-image super-resolution,'' \emph{Computer Vision and Image Understanding}, vol. 158, pp. 1--16, 2017.

\bibitem{ignatov2018pirm}
A.~Ignatov, R.~Timofte, T.~Van~Vu, T.~Minh~Luu, T.~X~Pham, C.~Van~Nguyen, Y.~Kim, J.-S. Choi, M.~Kim, J.~Huang \emph{et~al.}, ``Pirm challenge on perceptual image enhancement on smartphones: Report,'' in \emph{Proceedings of the European Conference on Computer Vision (ECCV) Workshops}, 2018, pp. 0--0.

\end{thebibliography}

\end{document}